\documentclass[10pt,twocolumn,letterpaper]{article}

\usepackage{times}
\usepackage{epsfig}
\usepackage{graphicx}
\usepackage{amsmath}
\usepackage{amssymb}
\usepackage[flushleft]{threeparttable}
\usepackage{caption}
\usepackage{fixltx2e}
\usepackage{enumitem}
\usepackage{marvosym}
\setlist[itemize]{noitemsep, nolistsep}

\usepackage{pifont}
\newcommand{\xmark}{\ding{55}}%

\usepackage{tabularx}
\usepackage{multirow}
\newcolumntype{?}{!{\vrule width 1pt}}
\newcolumntype{M}{>{$}c<{$}}
\newcolumntype{Z}{>{\centering\arraybackslash}X}
\newcolumntype{C}{>{\centering\arraybackslash}p}
\newcolumntype{Y}{>{\centering\arraybackslash}X}
\newcolumntype{L}{>{\raggedright\arraybackslash}p}
\newcommand{\specialcell}[2][c]{%
  \begin{tabular}[#1]{@{}c@{}}#2\end{tabular}}

\begin{document}

\title{Child Palm-ID: Contactless Palmprint Recognition for Children}

\author{Akash Godbole\textsuperscript{1}, Steven A. Grosz\textsuperscript{2} and Anil K. Jain\textsuperscript{3}, \textit{Life Fellow, IEEE}\\
Michigan State University\\
{\tt \{\textsuperscript{1}godbole1, \textsuperscript{2}groszste, \textsuperscript{3}jain\}@cse.msu.edu}
}

\maketitle
\thispagestyle{empty}

\begin{abstract}\vspace{-0.3cm}
Effective distribution of nutritional and healthcare aid for children, particularly infants and toddlers, in some of the least developed and most impoverished countries of the world, is a major problem due to the lack of reliable identification documents. Biometric authentication technology has been investigated to address child recognition in the absence of reliable ID documents. We present a mobile-based contactless palmprint recognition system, called Child Palm-ID, which meets the requirements of usability, hygiene, cost, and accuracy for child recognition. Using a contactless child palmprint database, Child-PalmDB1, consisting of 19,158 images from 1,020 unique palms (in the age range of 6 mos. to 48 mos.), we report a TAR=94.11\% @ FAR=0.1\%. The proposed Child Palm-ID system is also able to recognize adults, achieving a TAR=99.4\% on the CASIA contactless palmprint database and a TAR=100\% on the COEP contactless adult palmprint database, both @ FAR=0.1\%. These accuracies are competitive with the SOTA provided by COTS systems. Despite these high accuracies, we show that the TAR for time-separated child-palmprints is only 78.1\% @ FAR=0.1\%.
\end{abstract}
\vspace{-0.7cm}

\section{Introduction}\vspace{-0.2cm}

\par In 2020, 22\% of the world's 680 million children \cite{world}, under the age of 5 years, were physically stunted due to malnourishment and lack of adequate medication\footnote{https://www.who.int/data/gho/data/themes/topics/joint-child-malnutrition-estimates-unicef-who-wb}. A majority of these children live in developing or least developed countries where healthcare facilities and other resources are scarce. To address this problem, many international organizations such as the World Health Organization (WHO)\footnote{https://www.afro.who.int/news/strategic-plan-reduce-malnutrition-africa-adopted-who-member-states}, Bill and Melinda Gates Foundation (BMGF)\footnote{https://www.gatesfoundation.org/our-work/programs/global-growth-and-opportunity/nutrition} and the World Food Programme (WFP)\footnote{https://www.wfp.org/nutrition} have made substantial efforts to reduce the rate of malnourishment as well as improve vaccination coverage among this vulnerable population. However, the lack secure government-issued identification makes it difficult to authenticate the recipient of the services and curtail the occurence of fraud.
\begin{figure}[t]
\begin{center}
\includegraphics[width=0.95\linewidth]{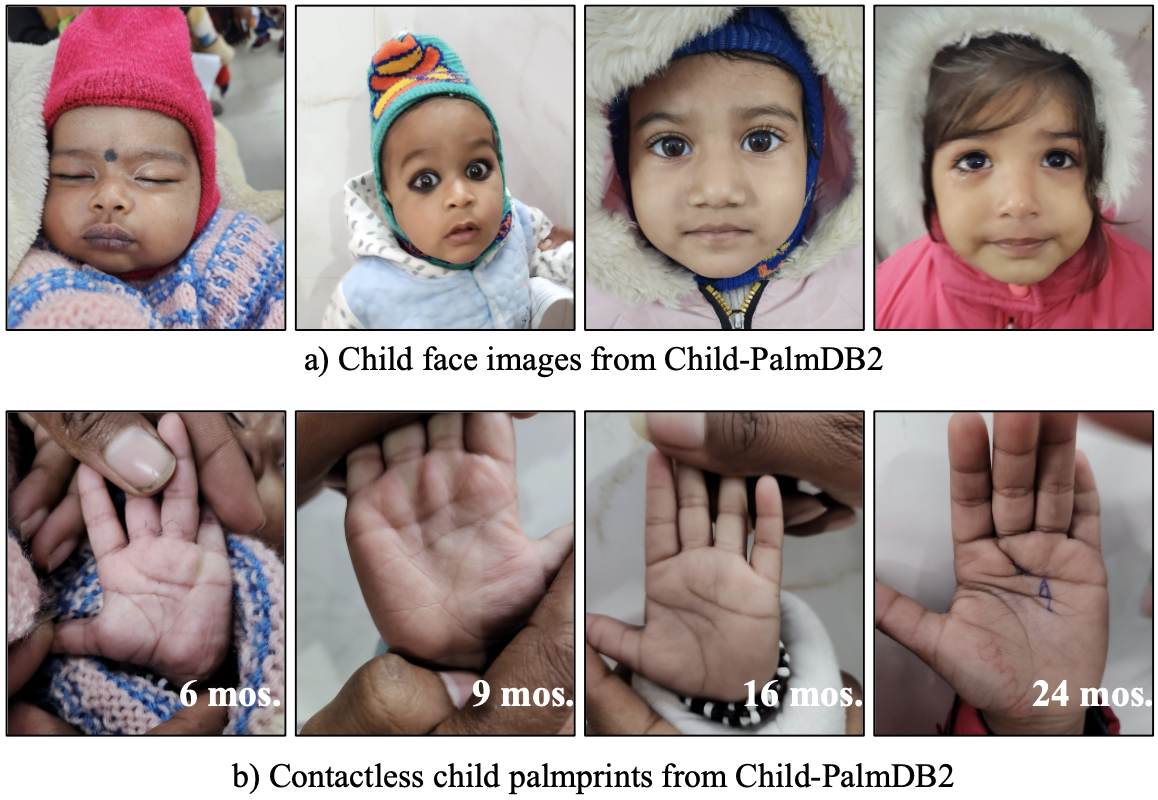}
\caption{\footnotesize Example face (a) and corresponding contactless palmprint images (b) in Child-PalmDB2 contactless palmprint database. }
\label{fig:face_collage}
\end{center}
\end{figure}

\begin{table*}[]

\caption{Summary of literature on biometric recognition of children}
\centering 
\begin{threeparttable}
\begin{tabular}{|C{0.08\linewidth}|C{0.1\linewidth}|C{0.155\linewidth}|C{0.17\linewidth}|C{0.18\linewidth}|C{0.17\linewidth}|}
\noalign{\hrule height 1.0pt}
\specialcell{\textbf{Authors}}   & \specialcell{\textbf{Modality}} & \specialcell{\textbf{Age Group}\\\textbf{(\# Subjects)}} & \specialcell{\textbf{Sensor}} & \specialcell{\textbf{Accuracy\tnote{\textbf{*}}}} & \specialcell{\textbf{Limitations}}\\
\noalign{\hrule height 1.0pt}
Jain et al. \cite{jain2016fingerprint}  & Fingerprint & 0-5 yrs (309) & Contact-based commercial and custom sensors & TAR=95\% \textbf{@} FAR=0.1\% using undisclosed matcher  & Recognition algorithm unknown\\
\noalign{\hrule height 0.5pt}
Saggese et al. \cite{saggese2019biometric} & Fingerprint & 0-18 mos. (504) & Custom contactless sensor & TAR=96.2\% \textbf{@} FAR=0.1\% using Verifinger\tnote{\textbf{**}}  & Complex fingerprint reader design\\
\noalign{\hrule height 0.5pt}
Engelsma et al. \cite{engelsma2021infant}  & Fingerprint & 8-16 weeks (315) & Custom contactless sensor & TAR=92.8\% \textbf{@} FAR=0.1\% using in-house matcher & Slow data acquisition \\
\noalign{\hrule height 0.5pt}
Kalisky et al. \cite{kalisky2022biometric}  & Fingerprint & 0-329 days (494) & Custom contactless sensor & TAR=77.0\% \textbf{@} FAR=0.1\% using Verifinger\tnote{\textbf{**}} & Low time-separated accuracy\\
\noalign{\hrule height 1.5pt}
Liu \cite{liu2017infant}  & Footprint & 1-9 mos. (60) & Contact-based commercial sensor & TAR=60\% \textbf{@} FAR=0.01\% using in-house matcher & Train and test set from same acquisition\\
\noalign{\hrule height 0.5pt}
Kotzerke et al. \cite{kotzerke2019newborn}   & Footprint & 2 days-6 mos. (60) & DSLR Camera & EER=22.22\% using in-house matcher & Lack of high quality data\\
\noalign{\hrule height 0.5pt}
Yambay et al. \cite{yambay2019feasibility}  & Toe print & 4-13 years (177) & Commercial contact-based sensor & EER=2.5\% using Verifinger\tnote{\textbf{**}} & Larger age group\\
\noalign{\hrule height 1.5pt}
Uhl and Wild \cite{uhl2009comparing}  & Palmprint & 3 yrs-18 yrs (301) & Flatbed scanner\tnote{\textbf{1}} & EER=4.63\% using in-house matcher  & Larger age group\\
\noalign{\hrule height 0.5pt}
Ramacha-ndra et al. \cite{ramachandra2018verifying}  & Contactless Palmprint & 6-36 hours (50) & Smartphone camera & EER=0.31\% using pre-trained AlexNet & Insufficient data for training and testing\\
\noalign{\hrule height 0.5pt}
Rajaram et al. \cite{rajaram2022palmnet}  & Contactless Palmprint & 3 mos-8 yrs (100) & Smartphone camera & EER=0.02\% using in-house matcher  & Train and test set from same acquisition\\
\noalign{\hrule height 0.5pt}
This paper  & Contactless Palmprint & 6 mos. - 4 yrs (515) & Smartphone camera & TAR=94.11\% \textbf{@} FAR=0.1\% & Low time-separated accuracy\\
\noalign{\hrule height 0.5pt}

\end{tabular}
\begin{tablenotes}
    \footnotesize
    \item[\textbf{*}] \textit{The studies listed above have used different evaluation metrics. Specifically, child fingerprint recognition and footprint recognition studies report TAR(\%) at FAR = 0.1\% (0.01\%) and the studies on toe prints and contactless palmprints report EER.}
    \item[\textbf{**}]\textit{ Version number unknown.}
    \item[\textbf{1}] \textit{The hand was placed at a stand-off above the flat-bed.}
\end{tablenotes}
\end{threeparttable}
\label{table:previous_work}
\end{table*}

\par Biometrics has received significant attention for the identification of children. However, biometrics-based identification solutions for children have yet to meet the requirements for field deployment, namely i) low-cost acquisition, ii) high accuracy, iii) robustness to capture environment (e.g. dust, humidity, and temperature), and iv) large throughput. Indeed, large-scale biometric identification systems in use today were not designed for use by very young children (infants and toddlers)\footnote{According to the Center for Disease Control (CDC), infants are between the ages of 0-1 yrs. and toddlers are between 2-3 yrs. https://www.cdc.gov/ncbddd/childdevelopment/positiveparenting/index.html}. The largest civil biometric system in the world, Aadhaar, only enrolls Indian residents over the age of 5 \cite{uidai}. This leaves a population of almost 118 million young children unaccounted for in India alone.

\par In addition to the above requirements for child biometric recognition, it is important to note that a biometric trait must meet the \textit{persistence} and \textit{individuality} requirements for the population under consideration \cite{jainbiometrics2011}. 
These requirements make it difficult to justify using face biometric since a child's face (both appearance and shape) changes significantly during the first few years after birth. A few studies have suggested using footprints and toe prints, but they neither satisfy the real-time acquisition requirement nor the ergonomics.
Iris images are difficult to capture if the child is sleeping or crying. Further, capturing iris may require the operator to forcibly open the child's eye which may make the parents uncomfortable. 
These limitations, paired with the rise of global virus outbreaks and concerns about hygiene, has motivated a push to develop biomteric systems that do not require physical contact with any capture surface \cite{amazon, fujitsu, redrock}. To satisfy all these requirements, we propose contactless palmprints as a cost-effective and robust solution for child identification. The proposed Child Palm-ID does not even require custom sensors, as in the case of fingerprint, footprint and iris since smartphone cameras have sufficient resolution to capture contactless palmprint images of children.

\par Table \ref{table:previous_work} shows some of the more prominent studies on biometric recognition for children. The fingerprint modality has been the popular choice thus far but recent studies have shown a trend towards contactless palmprints. The primary obstacle in contactless palmprint recognition for children is lack of training and evaluation data, both in terms of number of unique identities and longitudinal (time-separated) collections. Therefore, as part of this study, we collect three new datasets containing over 60,000 images from 1,824 unique child palms and 1,227 unique adult palms, called Child-PalmDB1 (August 2022), Child-PalmDB2 (January 2023) and Adult-PalmDB2 (January 2023), respectively (Fig. \ref{fig:data_collection}). Child-PalmDB1 and Child-PalmDB2 contain 159 common subjects (318 palms) for time-separated verification that we refer to as Child CrossDB.

\begin{figure*}[t]
\begin{center}
\includegraphics[width=0.95\linewidth]{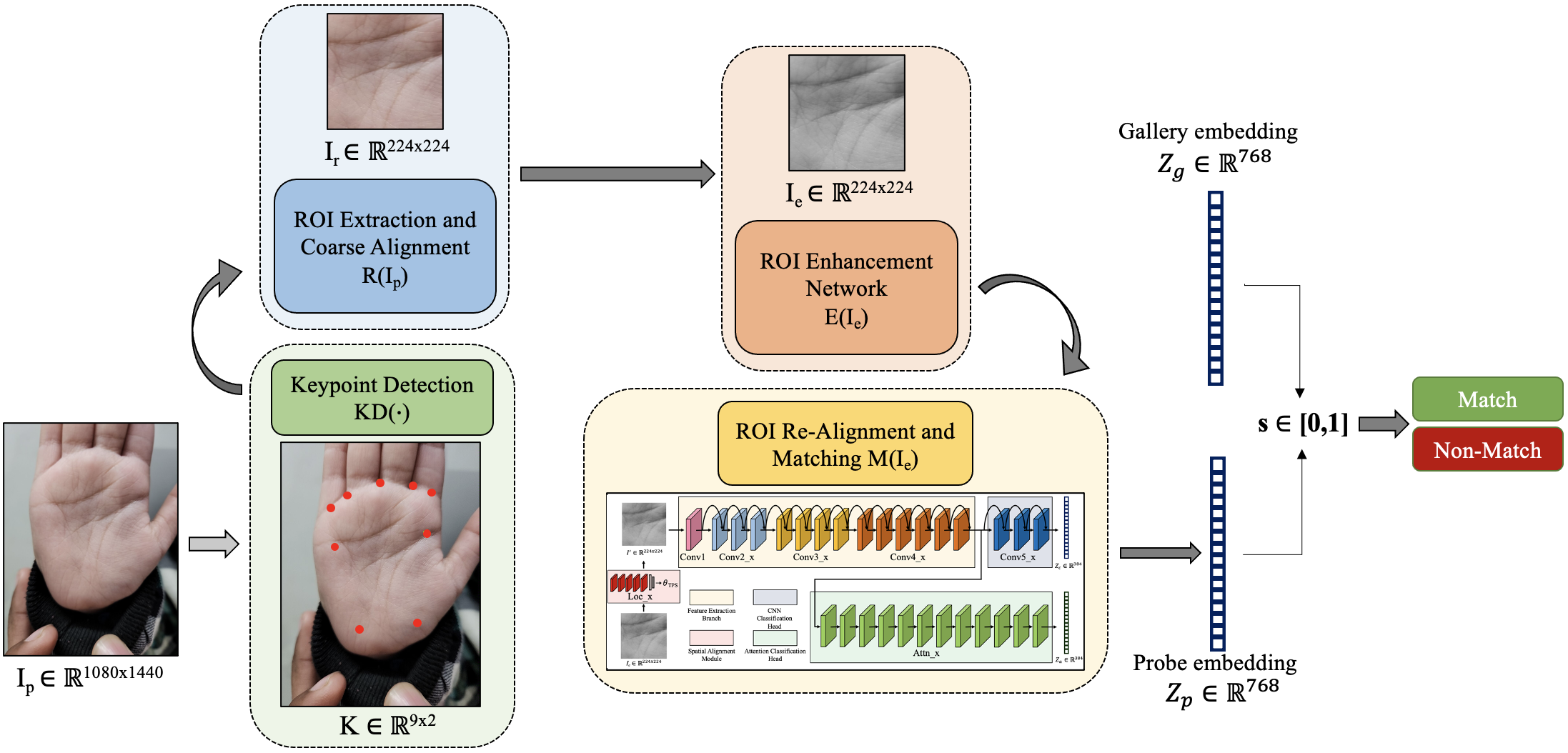}
\caption{\footnotesize A schematic diagram of the operational pipeline of Child Palm-ID. The input image $I_p$ is passed to the keypoint detection network $KD(\cdot)$. The coarse alignment between the probe and gallery images is based on a homographic transformation, followed by the AFR-Net architecture with a TPS unwarping module. The AFR-Net architecture diagram is adapted from \cite{grosz2022afr}.}
\label{fig:schematic}
\end{center}
\end{figure*}

\par Prior attempts at palmprint-based recognition for children focused on newborns and infants (less than 12 mos. old). These studies faced a number of challenges in palmprint capture of ``uncooperative'' subjects
\cite{lemes2011biometric}. To keep the child recognition problem tractable, we focus on children between 6 mos. to 48 mos. old. Child development studies \cite{dosman2012evidence} report that starting at the age of 12 mos., a child can follow instructions such as opening the fist and holding the palm in front of a mobile phone camera. This age group is also of interest to Aadhaar 2.0 \cite{press_info}, where one of the objectives is to lower the enrolment age which has been set at 5 yrs. since the inception of the program in 2009.\vspace{-0.05cm}
\par A contactless palmprint recognition system demands robustness to intra-class variability due to pose variations in palmprint images. The proposed Child Palm-ID addresses this problem by predicting landmarks on palm images coupled with a re-alignment of Regions of Interest (ROI) via a Thin Plate Spline (TPS) re-alignment module. Concretely, the contributions of this study are as follows:
\begin{itemize}
    \item A mobile-based contactless palmprint recognition system, Child Palm-ID, designed and prototyped for infants and toddlers.
    \item Keypoint detection and TPS re-alignment modules to handle large non-linear distortion and pose variations.
    \item Collection of Child-PalmDB1 and Child-PalmDB2 containing 1,824 unique child palms, and Adult-PalmDB2 containing 1,227 unique adult palms. These databases will be released once this paper is accepted for publication.
    \item Longitudinal contactless palmprint verification on Child CrossDB, a time-separated ($\sim$5 mos.) contactless child palmprint database containing 12,720 images from 318 unique palms.
\end{itemize}

\vspace{-0.4cm}
\section{Related Work}\vspace{-0.2cm}

\begin{figure*}[t]
\begin{center}
\includegraphics[width=0.97\linewidth]{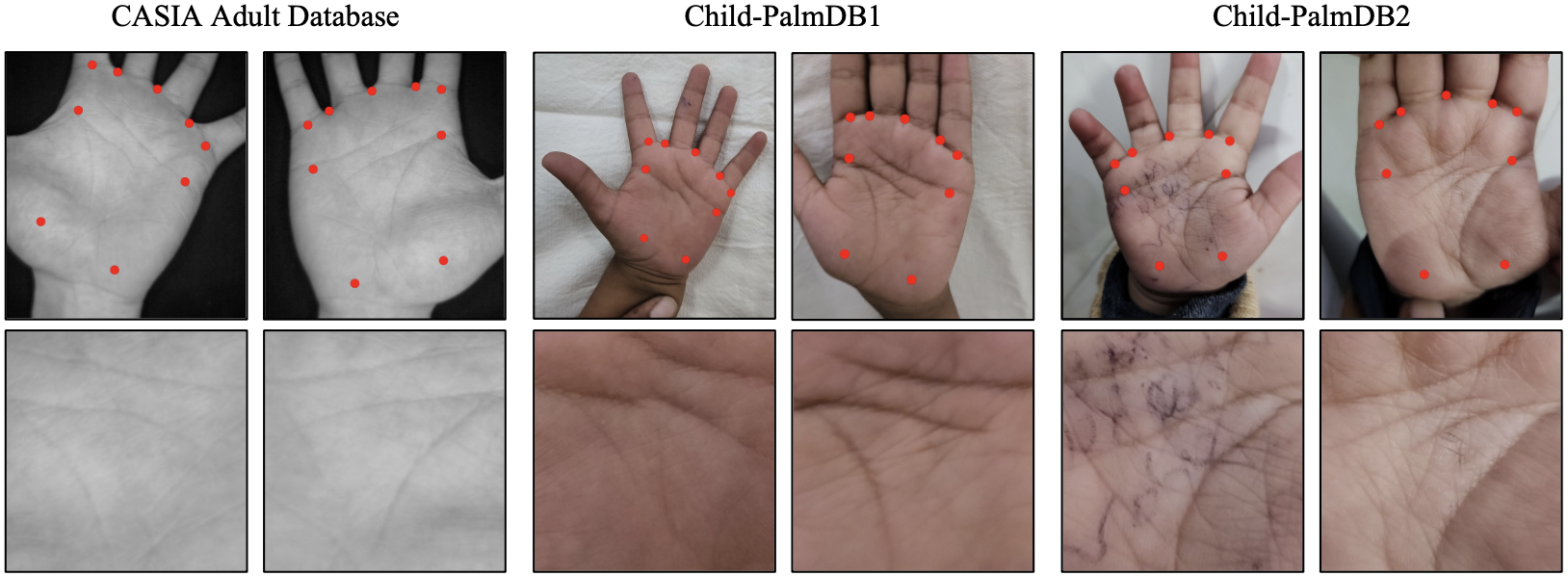}
\caption{\footnotesize Nine predicted keypoints overlaid (shown in red, top row) on the palmprint images from the CASIA Palmprint Image Database \cite{sunordinal}, Child-PalmDB1 and Child-PalmDB2. The bottom row shows the ROIs extracted via homographic transformation.}\vspace{-0.8cm}
\label{fig:keypoint_overlay}
\end{center}
\end{figure*}

\begin{figure}[t]
\begin{center}
\includegraphics[width=0.9\linewidth]{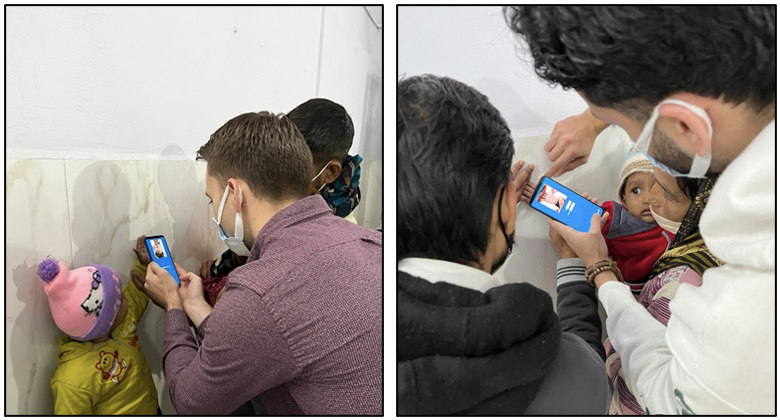}
\caption{\footnotesize Data collection camp in Dayalbagh, India, January 2023. The authors are collecting palmprint images using the PalmMobile SDK \cite{armatura}.}\vspace{-0.9cm}
\label{fig:data_collection}
\end{center}
\end{figure}

\par Contactless palmprint recognition consists of the following stages: i) Region of Interest (ROI) extraction, ii) ROI alignment and enhancement and iii) ROI comparison. See Fig. \ref{fig:schematic}. The predominant effort in the literature has been in building palmprint recognition systems for adults \cite{dian2016contactless, zhang2017towards, liu2020contactless, morales2011towards, wu2014sift, leng2017dual, liu2013coarse, jain2008latent, duta2002matching} while the focus on children has been limited \cite{rajaram2022palmnet, ramachandra2018verifying}. 

\vspace{-0.2cm}
\subsection{ROI Extraction}\vspace{-0.2cm}
\par Due to the potential of large pose variation in contactless palmprint image acquisition, it is important to obtain a consistent region of interest (ROI) across all the captured images. 

\par Depending on the nature of palmprint image acquisition, the ROI extraction method may vary. Handcrafted methods include binarizing the image to detect the finger valleys and using them to locate a square region on the palmar surface \cite{zhang2017towards}. This method may fail, for instance, if the fingers of the hand are not fully extended and separated from each other.
\par The acquisition of child palmprints may not always adhere to the above pose constraint requirement due to the continual development of fine motor control children have on their hands and fingers. Therefore, the proposed Child Palm-ID uses a deep-learning approach to predict a set of \textit{landmarks} to localize the ROI via a homographic transformation, an approach commonly used in face recognition \cite{wu2019facial, zhang2014facial} with larger pose variations.

\par We consider this landmark-based ROI to be \textit{coarsely aligned}, meaning it may require re-alignment for an accurate comparison with ROIs extracted from other palms. This will be further elaborated in sections 4.1 and 4.2.

\vspace{-0.2cm}
\subsection{ROI Alignment}\vspace{-0.2cm}

\par Adult palmprint recognition systems have utilized the principal lines, also referred to as palmar creases \cite{stevens1988development, kimura1986embryological}, for the re-alignment of ROIs \cite{wu2004palmprint, zhang2003online}. This method is effective provided that the palm capture adheres to pose constraints mentioned earlier. 

%

\par Recent advances in fingerprint and face recognition have turned to the use of Spatial Transformer Networks (STN), to predict alignment parameters that maximize the recognition accuracy \cite{ zhang2020end, engelsma2019learning, grosz2022afr}. Additionally, fine-tuned, non-linear alignment using a Thin Plate Spline (TPS) STN has shown even higher recognition performance in more unconstrained scenarios such as 3D facial recognition and contact-to-contactless fingerprint matching \cite{bhagavatula2017faster,grosz2021c2cl}.
\par In this paper, we implement a semi-supervised TPS STN module that learns a non-linear distortion field for a coarsely aligned ROI for improved accuracy of Child Palm-ID. 
\begin{figure}[t]
\begin{center}
\includegraphics[width=0.8\linewidth]{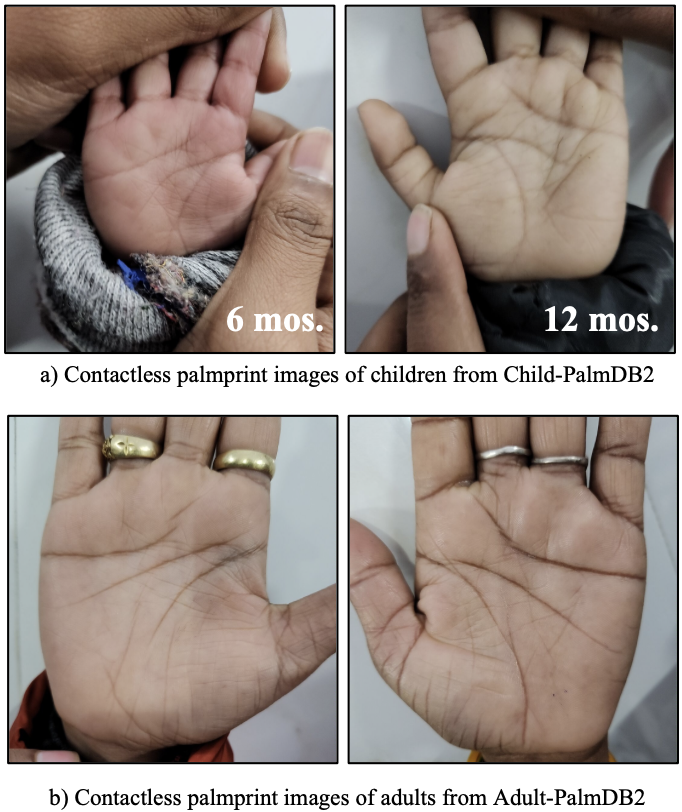}
\caption{\footnotesize Sample contactless palmprint images from (a) Child-PalmDB2 and (b) Adult-PalmDB2. For the child palmprint images, the age of the child is also included.}\vspace{-0.9cm}
\label{fig:jan_collection}
\end{center}
\end{figure}
\vspace{-0.25cm}
\subsection{ROI Matching}\vspace{-0.2cm}
\par Methods of comparing two palmprint images range from mathematical operations such as Fourier Transforms \cite{li2002palmprint} to learned embeddings from deep networks. A large proportion of recent studies have utilized deep networks to achieve compact embeddings (template) for high throughput, high accuracy and robustness compared to many handcrafted features \cite{dian2016contactless, jalali2015deformation, liu2018deep, liang2021compnet}. 

\vspace{-0.3cm}
\section{Databases}\vspace{-0.2cm}

\par We use a number of adult contactless palmprint databases available in the public domain, namely Tongji Adult Palmprint Database \cite{zhang2017towards}, CASIA Adult Palmprint Database \cite{sunordinal}
, CASIA Multispectral Database \cite{hao2008multispectral}
, COEP Adult Palmprint database \cite{coep}\footnote{ The COEP database contains 17 identities (136 images) that are mislabelled, which were excluded from the evaluation results.} and the Sapienza University Mobile Palmprint Datbase (SMPD) \cite{izadpanahkakhk2019novel}. We also utilize a private contactless palmprint video database containing 25-second video clips of 1,016 unique palms\footnote{We are unable to disclose details of this database due to NDA.}. These databases were collected using different capture devices and different image resolutions (from 600x800 px. to 3264x2448 px.). 
\begin{table}[h]

    \centering
    \begin{threeparttable}
    \caption{\vspace{-0.2cm}Details of databases used in this study}
        
            \begin{tabular}{|C{0.45\linewidth} | C{0.2\linewidth}|C{0.16\linewidth}|} 
                 \hline
                 \textbf{Training Database\tnote{\textbf{*}}} & \textbf{\# Unique Palms} & \textbf{Total \# images} \\ [1.0ex]
                 \noalign{\hrule height 1.2pt}
                    Tongji Adult \cite{zhang2017towards} &  600 & 12,000\\ 
                 \hline 
                    CASIA Multispectral \cite{hao2008multispectral} &  200 & 7,200\\
                 \hline 
                    Child-PalmDB2\tnote{\textbf{1}}  &  1,122 & 18,277\\
                 \hline
                    Adult-PalmDB2\tnote{\textbf{1}}  &  1,227 & 22,548\\
                 \hline
                    SMPD \cite{izadpanahkakhk2019novel} \tnote{\textbf{†}} &  92 & 3,677\\
                \hline
                    Private Database \tnote{\textbf{3}}&  1,016 & 28,748\\
                 \noalign{\hrule height 1.0pt}
                 \textbf{Testing Database} & \textbf{\# Unique Palms} & \textbf{Total \# images} \\
                 \hline
                    CASIA Adult \cite{sunordinal} &  614 & 5,502\\
                 \hline
                    COEP Adult \cite{coep} &  168 & 1,344\\
                 \hline
                    Child-PalmDB1\tnote{\textbf{2}} &  1,020 & 19,158\\
                \hline
                    Child CrossDB &  318 & 12,720\\
                 \noalign{\hrule height 1.2pt}
                 \hline
            \end{tabular}
        \begin{tablenotes}
            \footnotesize
            \item[\textbf{1}] \textit{Collected by authors. Will be released once the paper is accepted for publication.}
            \item[\textbf{2}] \textit{Collected by authors. Already in the public domain but anonymized for blind review.}
            \item[\textbf{3}] \textit{We extract individual frames from the video clip of each unique palm.}
            \item[\textbf{*}] \textit{Training and testing databases are disjoint.}
            \item[\textbf{†}] \textit{{https://www.kaggle.com/datasets/mahdieizadpanah/sapienza-university-mobile-palmprint-databasesmpd}}
        \end{tablenotes}\vspace{-0.5cm}
        \label{tab:datasets}
         \end{threeparttable}
    
\end{table}

\par The ages of the children in Child-PalmDB1 and Child-PalmDB2 range from a minimum of 6 mos. to a maximum of 48 mos. Child-PalmDB1 was collected in August 2022 whereas Child-PalmDB2 and Adult-PalmDB2 were collected in January 2023\footnote{Child-PalmDB1, Child-PalmDB2 and Adult-PalmDB2 were collected at Saran Ashram Hospital in Dayalbagh, India under approved IRB from both the hospital and the authors' institution.}. There are 159 overlapping subjects (excluded from the training set) in Child-PalmDB1 and Child-PalmDB2, called Child CrossDB, providing an avenue to explore time-separated contactless palmprint recognition performance for children. The palmprint images were collected using the Armatura PalmMobile SDK android application installed on a Samsung Galaxy S22 \cite{armatura}. The palm images for each child were collected with variations in roll, pitch, and yaw to best simulate a real world collection scenario. 
\par Table 2 shows the number of subjects and number of images in each of the databases\footnote{The authors are aware of PolyU-IITD and IITD Touchless Palmprint Database but despite our repeated requests, we were unable to obtain access to them.} used in this study as well as training and evaluation datasets. For the age groups of 6-12 mos., 12-24 mos. and 24-48 mos., Child-PalmDB1 contains 73, 161 and 230 subjects\footnote{Age information in Child-PalmDB1 is available for 444 subjects out of 515 subjects.}, respectively and Child-PalmDB2 contains 105, 202 and 375 subjects, respectively. For each subject, we collect images from both palms. So, the total number of unique palms is 2,142, including the 318 common palms between the two databases. 





\begin{figure}[t]
\begin{center}
\includegraphics[width=0.9\linewidth]{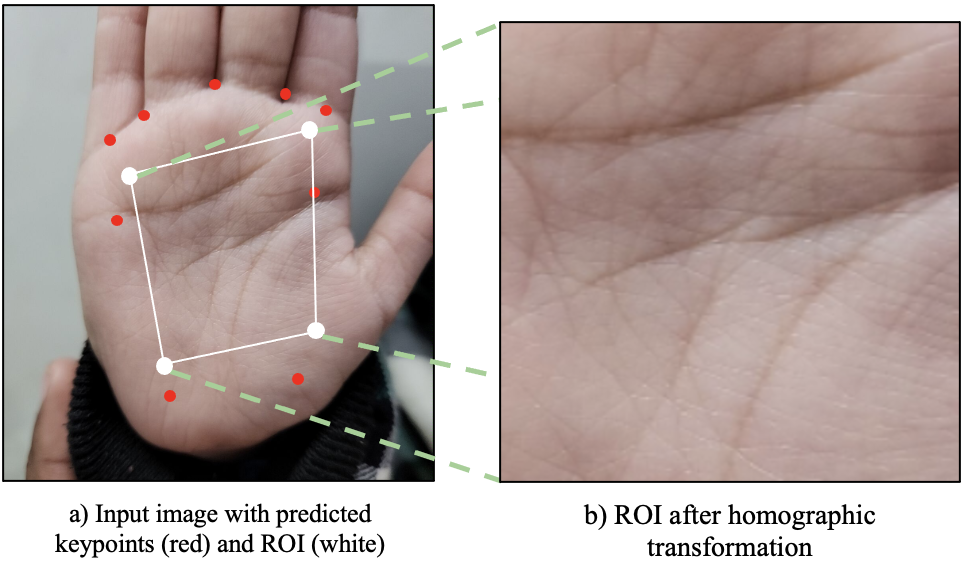}
\caption{\footnotesize Extracted keypoints (a) along with the ROI (b) after the coarse alignment. The polygon in (a) is an inverse homographic transformation of the four image vertices of (b) on (a) and represents the palmar region captured in the ROI.}\vspace{-0.7cm}
\label{fig:roi_extraction}
\end{center}
\end{figure}
\vspace{-0.3cm}
\section{Child Palm-ID Framework}\vspace{-0.2cm}
\par The operational pipeline of Child Palm-ID can be divided into four major components: i) Keypoint detection, ii) ROI extraction, iii) ROI enhancement, and iv) ROI re-alignment and matching. Fig. \ref{fig:schematic} shows a high-level schematic of Child Palm-ID. 

\begin{figure}[t]
\begin{center}
\includegraphics[width=0.95\linewidth]{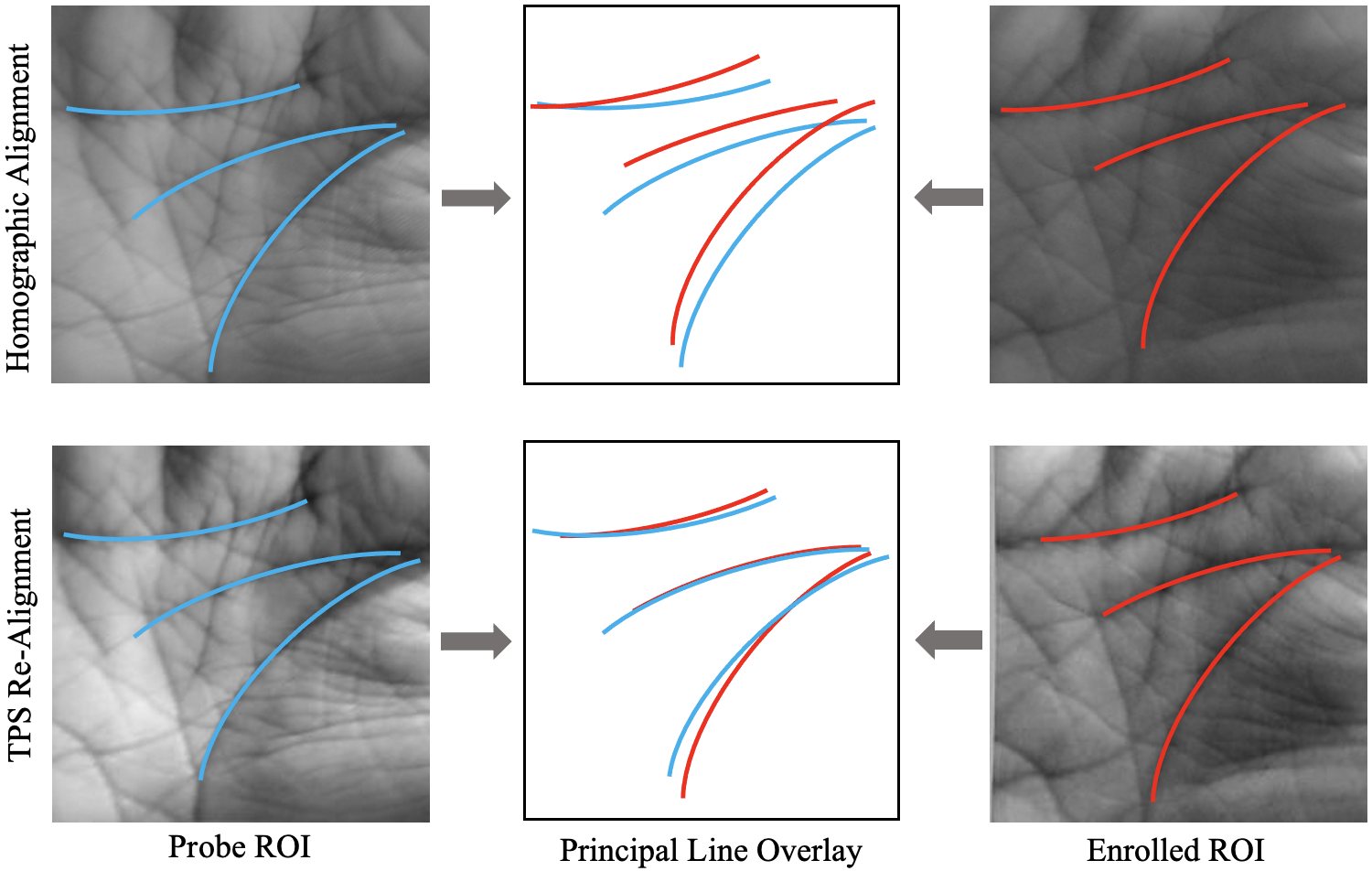}
\caption{\footnotesize Benefit of the TPS re-alignment. The ROIs in the bottom row are the re-aligned counterparts of the coarsely aligned ROIs in the top row.}\vspace{-0.8cm}
\label{fig:tps}
\end{center}
\end{figure}

\vspace{-0.2cm}
\subsection{Keypoint Detection}\vspace{-0.2cm}
\par The keypoint detection module $KD(\cdot)$ uses a ResNet-18 (see Table \ref{tab:datasets} for training set) architecture with two fully connected layers inserted at the end to predict 9 
keypoints \textit{K $\in \mathbb{R}^{9x2}$} in the input image ($I_p$); these keypoints are used to extract the coarsely aligned ROI. These 9 keypoints (fig \ref{fig:roi_extraction}) provide a degree of symmetry between the right and left hand while encompassing the the palm boundary containing salient information for a robust ROI. As groundtruth for training $KD(\cdot)$, we use the keypoints generated by the COTS system. An MSE objective function
is minimized to predict the keypoints. 
Fig. \ref{fig:keypoint_overlay} shows the predicted keypoints from $KD(\cdot)$ overlaid on the palmprint images on three databases. The predicted keypoints along with the input image serve as input to the ROI extraction module.

\vspace{-0.2cm}
\subsection{ROI extraction}\vspace{-0.2cm}
\par Accurate alignment of palmprint images to a consistent coordinate system is essential to extract a robust, albeit coarsely aligned, ROI. To homogenize the coordinate system, a set of 9 destination points \textit{D} are selected to perform a 9-point homographic transformation $H(\cdot)$ between $K$ and $D$ yielding a perspective transform matrix $\theta_h$. The ROI module $R(\cdot)$ applies $\theta_h$ to $I_p$ to get a warped image $I^w_p$; a 224 x 224 cropped image $C$ is extracted yielding the coarsely aligned ROI, $I_r$ (eqs. \ref{homo}, \ref{warp}, \ref{roi}).

\begin{equation}
    \theta_h = H(K, D)
\label{homo}
\end{equation}
\begin{equation}
    I^w_p = R(I_p;\theta_h)
\label{warp}
\end{equation}
\begin{equation}
    I_r = C(I^w_p, 224)
\label{roi}
\end{equation}


\begin{figure}
    \centering
    \includegraphics[width=0.95\linewidth]{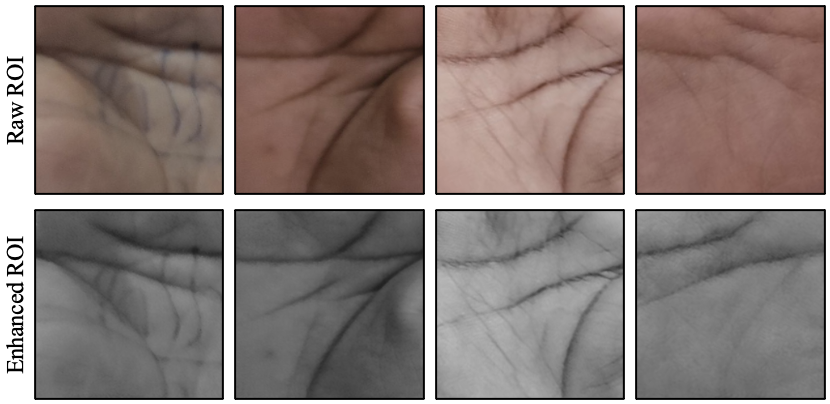}
    \caption{\footnotesize Extracted ROIs (top row) and the corresponding enhanced ROIs (bottom row).}\vspace{-0.6cm}
    \label{fig:enh}
\end{figure}
\vspace{-0.3cm}
\subsection{ROI enhancement}\vspace{-0.1cm}
\par We utilize the autoencoder network designed to enhance poor quality latent fingerprints in LFR-Net \cite{grosz2023latent} and adapt it to enhance contactless palmprints. Due to the unconstrained nature of contactless palm capture, many of the captured images exhibit very poor ridge contrast, motion blur, and other degradations. To help mitigate these challenges, for training the autoencoder we simulated low quality palm images by, for example, blurring and downsampling as data augmentations. The autoencoder is trained via an MSE loss between the high quality palmprint inputs and the reconstructed outputs of the enhancement network. Example enhanced images are shown in Fig. \ref{fig:enh} and the benefit of the enhancement network is shown quantitatively in the ablation study in section 5.2.
\vspace{-0.2cm}
\subsection{Re-Alignment and Matching}\vspace{-0.2cm}
\par The feature extraction and matching architecture of Child Palm-ID is based on AFR-Net \cite{grosz2022afr}, a fingerprint recognition model based on ResNet50 \cite{he2016deep} and Vision Transformers (ViT) \cite{dosovitskiy2020image}. AFR-Net uses an STN to predict an affine alignment of the input images. We modify the STN to predict a TPS alignment that applies a non-linear, learned, distortion field $\theta_{TPS}$ to the coarsely aligned palmprint ROIs ($I_r$) producing an aligned ROI, $I'$ (eq. \ref{tps_eqn}). 
\begin{equation}
    I' = T(I_r; \theta_{TPS})
\label{tps_eqn}
\end{equation}
\par A learned TPS network has been shown to boost performance in fingerprint and face matching \cite{grosz2021c2cl, bhagavatula2017faster}. 
Fig. \ref{fig:tps} shows the improved alignment between two ROIs after applying $T(I_r)$. Affirming the intuition behind the use of $T(\cdot)$, a significant boost in recognition performance was observed compared to the use of the pre-existing STN in AFR-Net (from TAR = 73.8\% to TAR = 88.3\%, both at FAR = 0.1\%. See Table \ref{tab:ablation}.).
\par The probe and gallery embeddings $Z_p$ and $Z_g$, respectively, are compared to obtain a similarity score $s\in [0, 1]$ (eq. \ref{similarity}).
\begin{equation}
    s = Z^T_p \cdot Z_g , \in [0, 1]
\label{similarity}
\end{equation}
\begin{figure}
    \centering
    \includegraphics[width=0.95\linewidth]{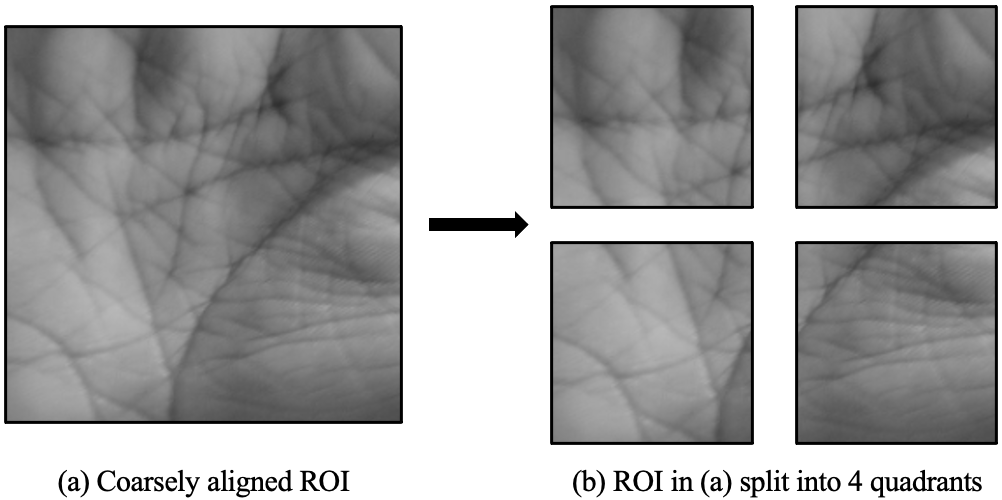}
    \caption{\footnotesize The coarsely aligned and enhanced ROI (a) and its 4 quadrants (b). A model is trained for each quadrant and the whole ROI. Final similarity score is a fusion of the 5 model scores. }\vspace{-0.3cm}
    \label{fig:score_fusion}
\end{figure}
\vspace{-0.9cm}
\subsubsection{Ensemble of multi-patch embeddings}\vspace{-0.2cm}
\par The crux of ensemble learning is utilizing multiple complimentary models that improve the overall performance of the system via different fusion techniques \cite{sagi2018ensemble, godbole2022learning}. We divide the 224x224 coarsely aligned, enhanced, ROIs into 4 quadrants (Fig. \ref{fig:score_fusion}) and train an ensemble of models, one per quadrant to complement the model trained on the entire ROI. Using the ensemble of these five embeddings, we obtain a final similarity score based on mean score fusion. We show the quantitative benefit of the ensemble in the ablation study (Table \ref{tab:ablation}).

\begin{figure}
    \centering
    \includegraphics[width=0.95\linewidth]{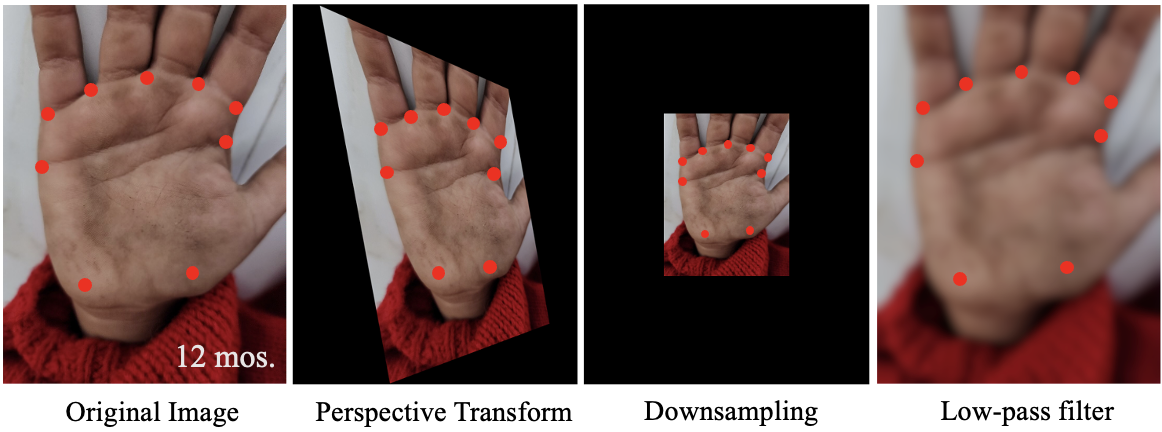}
    \caption{\footnotesize Examples of data augmentations for Child Palm-ID training. Rotation and translation augmentations are not shown.}\vspace{-0.5cm}
    \label{fig:augs}
\end{figure}
\vspace{-0.2cm}
\subsection{Training Details}\vspace{-0.2cm}
\par Child Palm-ID was trained using an ArcFace loss function with a margin of 1.5, learning rate of 1e-4, weight decay of 2e-5, and polynomial learning rate decay function with a power of 3 and minimum learning rate of 1e-5. All models were trained with a batch size of 64 on a Nvidia GeForce RTX 2080Ti GPUs for 75 epochs. Furthermore, some key data augmentations (translation, rotation, scaling, blurring and perspective transforms) were randomly applied during the training process to improve the accuracy of the recognition system on images with large pose variations, commonly observed in child palmprint images. Fig. \ref{fig:augs} shows examples of augmentations applied on a single palmprint image. Note that any number of these augmentations may be applied to a single image during training with a probability of $p = 0.5$.

\vspace{-0.3cm}
\section{Experimental Results}\vspace{-0.2cm}
\label{exp_results}
\par In this section we evaluate verification performance of Child Palm-ID and compare it to the baseline accuracy of the COTS system \cite{armatura}. We report the accuracy on the entire Child CrossDB as well as the three age subgroups (6-12 mos., 12-24 mos. and 24-48 mos.) from Child-PalmDB1. Finally, we conduct an ablation study to examine the effects of re-alignment, learned enhancement, ensemble of embeddings and data augmentations on the performance of Child Palm-ID.
\vspace{-0.2cm}
\subsection{Verification Results}\vspace{-0.2cm}

\par We report verification performance on four evaluation databases that were altogether kept separate from the training set (see Table \ref{tab:datasets}). The recognition performance of the proposed Child Palm-ID is competitive with COTS\footnote{The architecture and training set for the COTS is not known to us and both the adult databases used for evaluation are in the public domain.}. We also report the longitudinal verification performance on the Child CrossDB containing the 159 subjects present in both Child-PalmDB1 and Child-PalmDB2 in Table \ref{tab:tar}. Images in Child CrossDB were not included in the training set. 
\begin{table}[h]
    
    \centering
    
    \caption{TAR(\%) @ FAR=0.1\% of Child Palm-ID and COTS.\\}
    \begin{threeparttable}
        \begin{tabular}{|C{0.2\linewidth} | C{0.2\linewidth}|C{0.2\linewidth}|C{0.2\linewidth}|} 
             \hline
             \textbf{Database} & \textbf{Child Palm-ID} & \textbf{COTS \cite{armatura}}  & \textbf{Child Palm-ID + COTS}\\ [1.0ex]
             \noalign{\hrule height 1.2pt}
                CPDB1\tnote{\textbf{\dag}} (all ages) &  94.11 & 92.72 & \textbf{94.46}\\
             \hline
                CPDB1\tnote{\textbf{\dag}} (6-12 mos.) &  91.76 & 89.88 & \textbf{92.48}\\
             \hline
                CPDB1\tnote{\textbf{\dag}} (12-24 mos.) &  95.74 & 93.89 & \textbf{96.12}\\
            \hline
                CPDB1\tnote{\textbf{\dag}} (24-48 mos.) &  98.86 & 96.32 & \textbf{98.97}\\
             \hline
                Child CrossDB \tnote{\textbf{*}} &  78.1 & 78.22 & \textbf{82.02}\\
             \noalign{\hrule height 1.2pt}
                CASIA Adult &  99.4 & 100 & 100\\
             \hline
                COEP Adult\tnote{\textbf{\ddag}} &  100 & 100 & 100\\
             \noalign{\hrule height 2pt}
             \hline
        \end{tabular}
    
    \begin{tablenotes}
        \footnotesize
        \item[\textbf{\dag}]\textit{ We abbreviate Child-PalmDB1 as CPDB1 in this table to save space.}
        \item[\textbf{\ddag}] \textit{17 mislabelled identities were removed.}
        \item[\textbf{*}] \textit{Child CrossDB was not included in the training set.}
    \end{tablenotes}\vspace{-0.3cm}
    \end{threeparttable}
    
    \label{tab:tar}
\end{table}
\par
\begin{figure}
    \centering
    \includegraphics[width=0.95\linewidth]{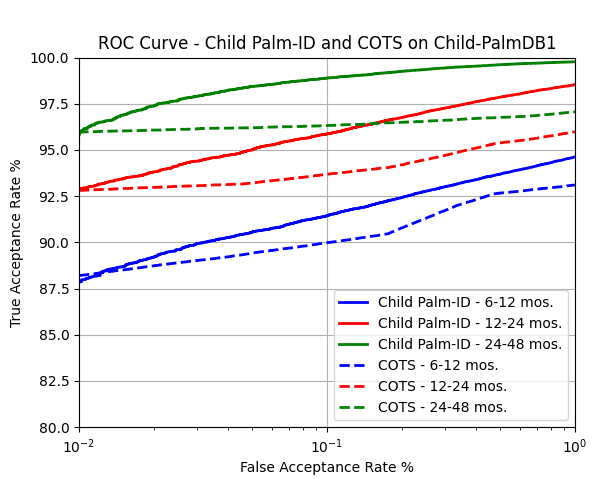}
    \caption{\footnotesize ROC curves comparing the performance of Child Palm-ID against the COTS system \cite{armatura} on Child-PalmDB1.}\vspace{-0.3cm}
    \label{fig:roc}
\end{figure}

 It is instructive to notice the trend in performance of Child Palm-ID on different age groups. Intuitively, a recognition system would perform better on relatively older children since they are likely to be more cooperative during data acquisition. Child Palm-ID shows an accuracy of TAR=91.76\% on children between the ages of 6 to 12 mos., TAR=95.74\% on children between the ages of 12 to 24 mos. and TAR=98.86\% on children in the age group of 24-48 mos., all @ FAR=0.1\%. Fig. \ref{fig:roc} shows that Child Palm-ID outperforms the COTS system at FAR = 0.1\% in each of the three evaluation age groups. We show an improvement by sum score fusion of Child Palm-ID and COTS, especially in the case of Child CrossDB at FAR=0.1\% (see supplementary material for more details).
\vspace{-0.2cm}
\subsection{Ablation Study}\vspace{-0.2cm}
\par In the ablation study, we examine the effects of the autoencoder enhancement module, TPS alignment module, multi-patch embeddings and data augmentations for training. The training datasets were fixed (Table \ref{tab:datasets}) in these ablations. The results of the ablation study are shown in Table \ref{tab:ablation}. The TPS re-alignment module in row 2 of Table \ref{tab:ablation}, gives the biggest boost in accuracy on all the four evaluation databases. The image enhancement, ensemble of embeddings and data augmentations further boost the accuracy.

\begin{table*}

    \centering
    \captionsetup{justification=centering}
    \caption{Ablation Study for Child Palm-ID. Results are reported as TAR (\%) @ FAR = 0.1\%}
    \begin{threeparttable}
    \begin{tabular}{| C{0.08\linewidth} | C{0.08\linewidth} |  C{0.09\linewidth} |  C{0.08\linewidth} |  C{0.08\linewidth} ||  C{0.10\linewidth} |  C{0.10\linewidth} |  C{0.10\linewidth}| C{0.10\linewidth}| C{0.10\linewidth}|} 
         \hline
          \multicolumn{5}{|c||}{\textbf{Modules Used}} & \multicolumn{4}{c|}{\textbf{Evaluation Databases}}  \\ [1.0ex]
         \noalign{\hrule height 1.2pt}
         Coarse Alignment & Re-Alignment & Data Augmentation & Ensemble of Embeddings & Enhanc-ement & Child-PalmDB1 & CASIA Adult Database & COEP Adult Database & Child CrossDB\\
         \noalign{\hrule height 1.2pt}
         \checkmark & \xmark & \xmark & \xmark & \xmark & 73.8 & 92.4 & 91.6 & 66.56\\
         \hline
         \checkmark & \checkmark & \xmark & \xmark & \xmark & 88.3 & 98.8 & 99.1 & 74.68\\
         \hline
         \checkmark & \checkmark & \checkmark & \xmark & \xmark & 92.43 & 99.1 & 100 & 76.67\\
         \hline
         \checkmark & \checkmark & \checkmark & \checkmark & \xmark & 93.01 & 99.6 & 100 & 77.4\\
         \hline
         \checkmark & \checkmark & \checkmark & \checkmark & \checkmark & 94.11 & 99.4 & 100 & 78.1\\
         \noalign{\hrule height 1.2pt}
    \end{tabular}\vspace{-0.5cm}
    \end{threeparttable}
    \label{tab:ablation}
\end{table*}

\begin{figure}
    \centering
    \includegraphics[width=0.95\linewidth]{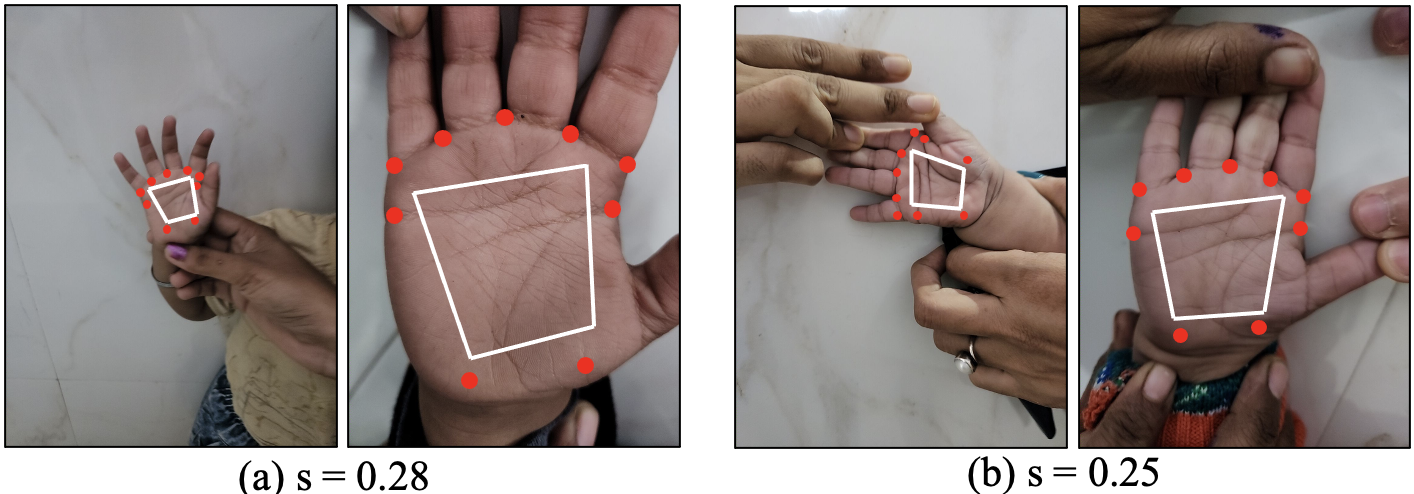}
    \caption{\footnotesize Failure cases of Child Palm-ID in Child CrossDB. For each genuine pair of images, the similarity score $s$ is below the threshold of 0.46 at FAR = 0.1\%. In both (a) and (b), the left image is from Child-PalmDB1 and the right image is from Child-PalmDB2.}\vspace{-0.3cm}
    \label{fig:crossdb_failures}
\end{figure}
\begin{figure}
    \centering
    \includegraphics[width=0.95\linewidth]{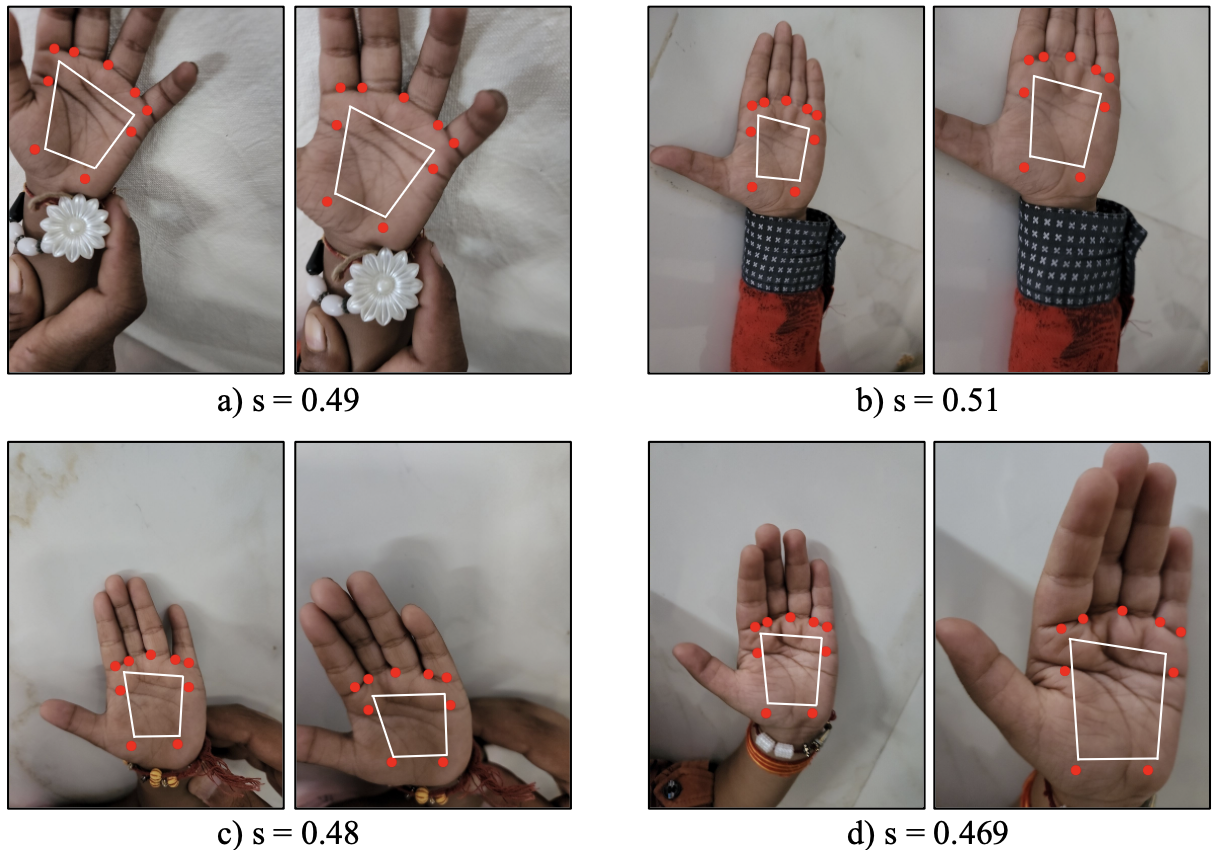}
    \caption{\footnotesize Successful cases of Child Palm-ID in Child-PalmDB1. For each genuine pair of images, the similarity score $s$ is above the threshold of 0.46 at FAR = 0.1\%.}\vspace{-0.38cm}
    \label{fig:success}
\end{figure}
\vspace{-0.2cm}
\subsection{Failure Cases}\vspace{-0.25cm}
\par
\begin{figure}
    \centering
    \includegraphics[width=0.95\linewidth]{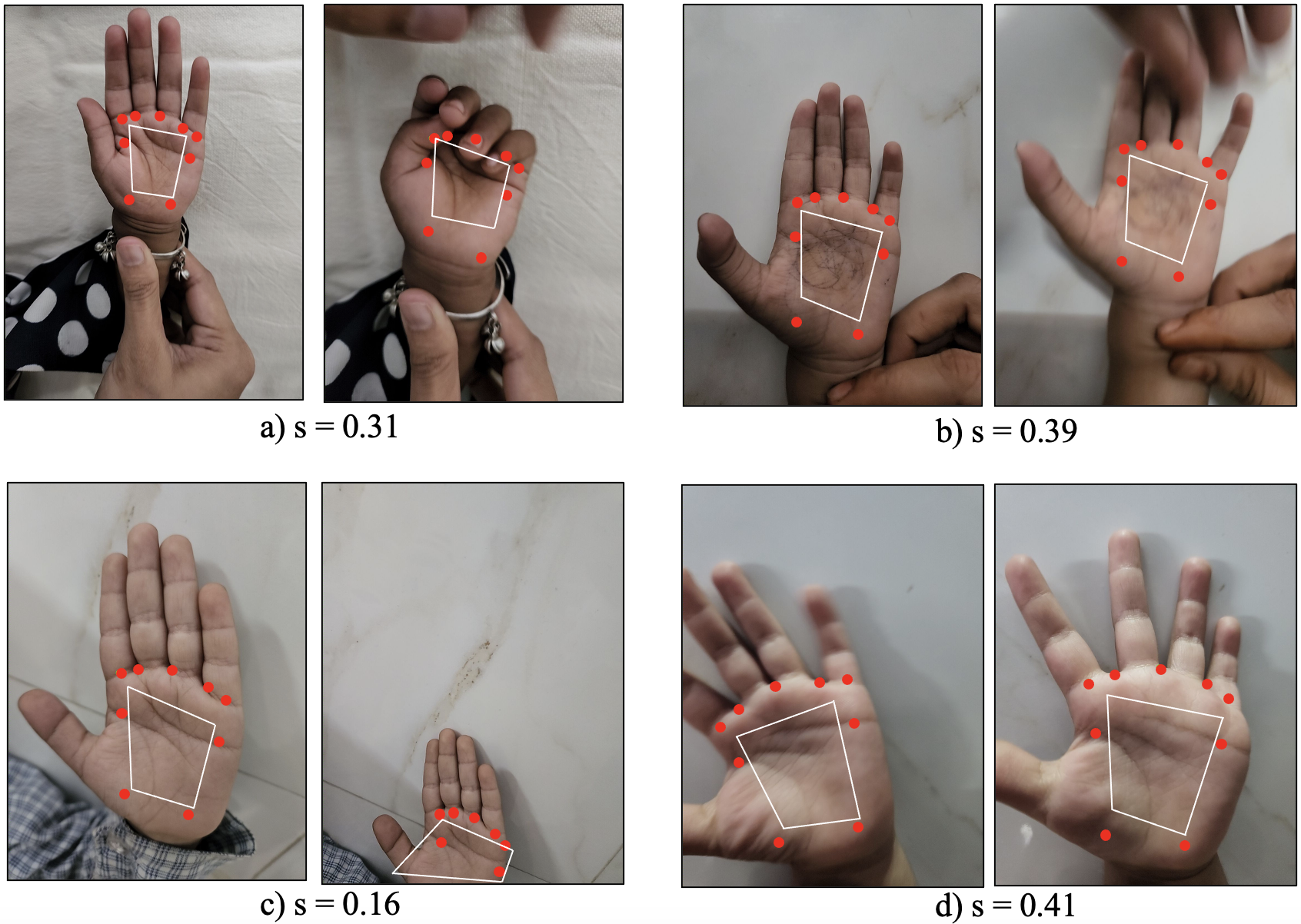}
    \caption{\footnotesize Failure cases of Child Palm-ID in Child-PalmDB1. For each genuine pair of images, the similarity score $s$ is below the threshold of 0.46 at FAR = 0.1\%.}\vspace{-0.5cm}
    \label{fig:failures}
\end{figure}

 Fig. \ref{fig:failures} shows the failure cases of Child Palm-ID when evaluated on Child-PalmDB1. The main reasons for failures are i) poor image quality and ii) severe variation in pose between two images. 
\par Figs. \ref{fig:failures}(b) and \ref{fig:failures}(d) indicate the poor quality images that mainly arise due to the unexpected movement of the child's palm during the image acquisition process. The highlighted ROIs include partially closed fingers and incorrectly detected keypoints due to interference from the background. This can be mitigated by either implementing a palmprint quality metric to filter out such images or with adult supervision during palmprint acquisition.
In Fig. \ref{fig:failures}(a), the subject's fist is partially closed and is partially occluded in Fig. \ref{fig:failures}(c). This leads to incorrect keypoint detection. However, Figs. \ref{fig:failures}(a) and \ref{fig:failures}(c) may also represent the genre of images that an untrained technician might acquire, thereby bearing some resemblance to an operational scenario.
\par Fig. \ref{fig:crossdb_failures} shows the failure cases of Child Palm-ID on Child CrossDB. This highlights the challenges in cross-dataset comparison where there are significant differences in standoff distance, lighting and rotation between two time-separated acquisitions.

\vspace{-0.2cm}
\subsection{Contactless child palmprint acquisition}\vspace{-0.2cm}
\par As outlined previously, collecting palmprint images of a child is a challenging exercise that requires carefully designed protocols. From our experience, we recommend the following crucial guidelines for collection of high quality data:
\begin{itemize}
    \item Adult supervision to prevent unexpected movements of the child's palm.
    \item Fixed range of standoff distances and pose variations.
    \item Consistent and uniform lighting to reduce shadows and maintain high contrast in the images.
\end{itemize}
\vspace{-0.3cm}
\section{Conclusion and Future Work}\vspace{-0.2cm}
\par Biometric recognition systems have made great strides over the past 20 years in terms of acquisition, accuracy, cost, and broad range of deployments ranging from mobile phone unlock to large-scale national ID. However, all these systems were designed to be used by adults. Yet there are numerous social good tasks ranging from eradicating vaccine preventable diseases to child malnutrition where biometric recognition can play a significant role to prevent misery and loss of life.
\par We have designed and prototyped Child Palm-ID, a contactless mobile-based palmprint recognition system geared towards children. We have evaluated verification performance of Child Palm-ID on both child as well as adult contactless palmprint databases. We show competitive recognition performance of our system as compared against a SOTA COTS system @ FAR=0.1\%. The main technical contributions of our paper include a re-alignment strategy for palmprint images using a TPS alignment module and an autoencoder-based image enhancement. Furthermore, we will place our database collection, two child and one adult contactless palmprint datbases in public domain once this paper has been accepted. Future work may include i) Child Palm-ID mobile app displaying the faces of the top $N$ retrievals from a gallery for a probe so the operator is able to manually confirm the identity of the child, ii) introduction of a palmprint image quality metric to filter images of poor quality, iii) multimodal biometric recognition for children, iv) synthetic palmprint generation to amplify the amount of data available for training.

{
\bibliography{egbib}
}

\section{Appendix}
\subsection{Score distributions of Child Palm-ID and the COTS system}
\par As mentioned in section 5.1 of the paper, the sum score fusion of the similarity scores from Child Palm-ID and the COTS matcher provided an additional improvement in accuracy across all evaluation databases. This score fusion was particularly important to show the potential for improvement in the case of Child CrossDB. We show that the performance on Child CrossDB improves from TAR=78.1\% (for Child Palm-ID) and TAR=78.22\% (for COTS) to TAR=82.02\% (Child Palm-ID + COTS), all at FAR=0.1\%. The contingency table of genuine and imposter comparisons for both matchers helped understand the potential benefit of fusing their results. Table 1 shows the contingency table for the genuine distribution of Child CrossDB. This shows the number of comparisons where i) both matchers gave the same decision (diagonal entries) and ii) both matchers gave different decisions (cross-diagonal entries). Each decision is binary in terms of match/non-match.
\begin{table}[h]
    
    \centering
    
    \caption{Contingency table of Child Palm-ID and COTS on Child CrossDB, pre-fusion.\\}
    \begin{threeparttable}
        \begin{tabular}{|C{0.1\linewidth} | C{0.25\linewidth}|C{0.2\linewidth}|C{0.2\linewidth}|} 
             \hline
              &  & \multicolumn{2}{c|}{\textbf{COTS}}\\ [1.0ex]
             \noalign{\hrule height 1.2pt}
                 &   & Match & Non-Match\\
             \hline
                \textbf{CPID}\tnote{\dag} &  Match & 162,696 & 10,805\\
             \hline
                \textbf{CPID}\tnote{\dag} &  Non-Match & 21,810 & 31,316\\
             \hline
        \end{tabular}
    
    \begin{tablenotes}
        \footnotesize
        \item[\textbf{\dag}]\textit{ We abbreviate Child Palm-ID as CPID in this table.}
    \end{tablenotes}\vspace{-0.3cm}
    \end{threeparttable}
    
    \label{tab:tar}
\end{table}

\par We see that the cross-diagonal elements, where the decisions of the matchers are different, are key to improving the performance after fusion. Ideally, both these numbers would be 0.
\par The architecture of the COTS system is unknown to us. However, we believe the Child Palm-ID system to be distinct from the COTS while showing competitive accuracy across all evaluations, motivating the fusion of the two. Fig. 1 in this document shows the genuine and impostor distributions of Child Palm-ID and the COTS before the fusion.
\par It appears that the COTS has higher score separation compared to Child Palm-ID. To fuse the scores, we multiply the Child Palm-ID scores by 100 to have the same score	range as the COTS ([0, 100]) and then we simply sum the two scores to obtain the fused result. Fig. 2 shows the score distribution after the fusion ([0, 200]).
\begin{figure}
    \centering
    \includegraphics[width=\linewidth]{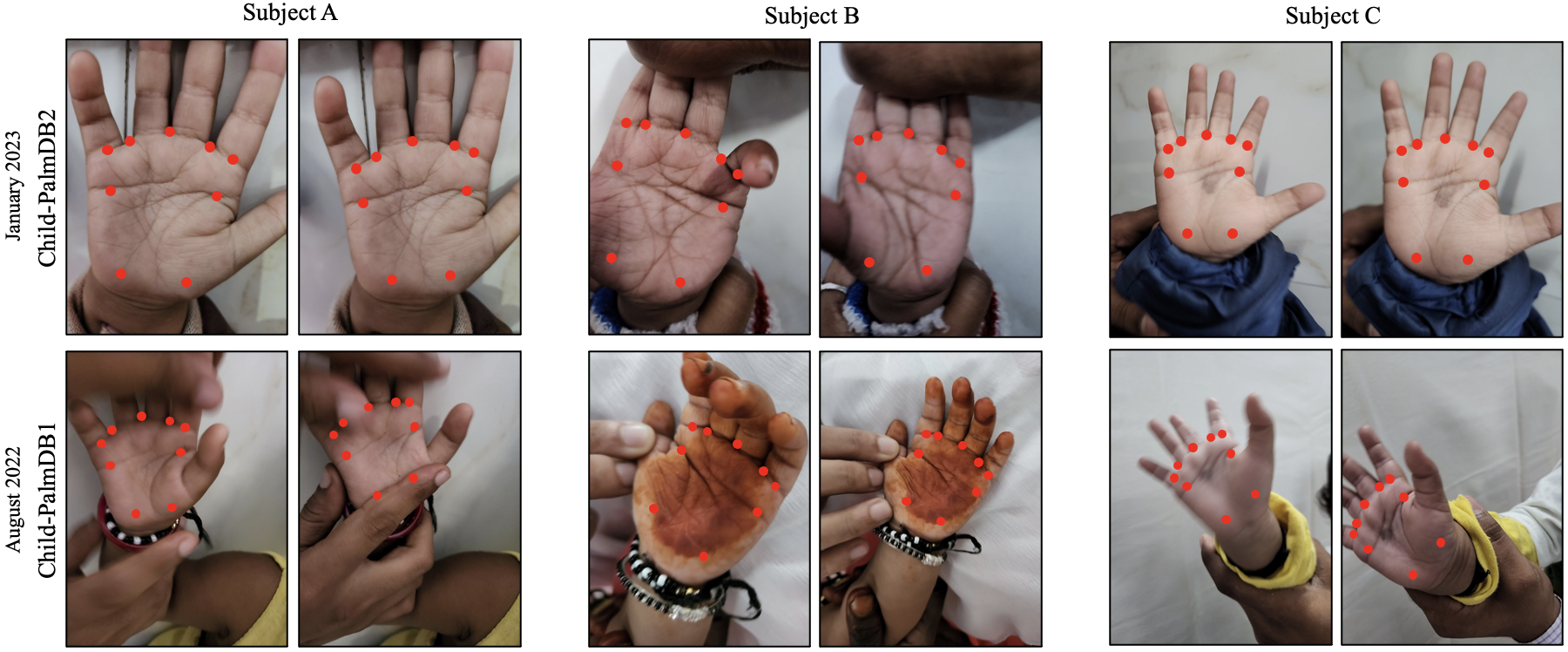}
    \caption{Example images of the subjects in Child CrossDB that are responsible for a majority of the errors}
    \label{fig:more_misclassified}
\end{figure}
\par We can see that visually, the separation between the genuine and imposter scores has	 improved compared to the individual distributions of either matcher in Fig. 1. Quantitatively, 	 
this results in a higher accuracy for Child CrossDB. We repeat the same fusion technique for all 	evaluation databases and observe an improvement for each of them at FAR=0.1\% (see Table 3.		in the paper).

\subsection{Additional Failure Cases in Child CrossDB}
\par The primary reason for the lower performance of both Child Palm-ID and the COTS is due to the underlying differences in the nature of the two child palmprint databases, Child-PalmDB1 and Child-PalmDB2. The images in Child-PalmDB1 contain a much larger variation in pose, higher standoff distances and lower image quality compared to Child-PalmDB2. We look at some of the subjects in Child CrossDB that cause most of the errors in classification in Fig. \ref{fig:more_misclassified}. The top row is images from Child-PalmDB2 and the bottom row is images from Child-PalmDB1. We can see that the causes of error range from large stand-off distances to the hand being covered in henna, etc.

\subsection{Subjects With Large Intra-Class Variability}
\begin{figure}
    \centering
    \includegraphics[width=\linewidth]{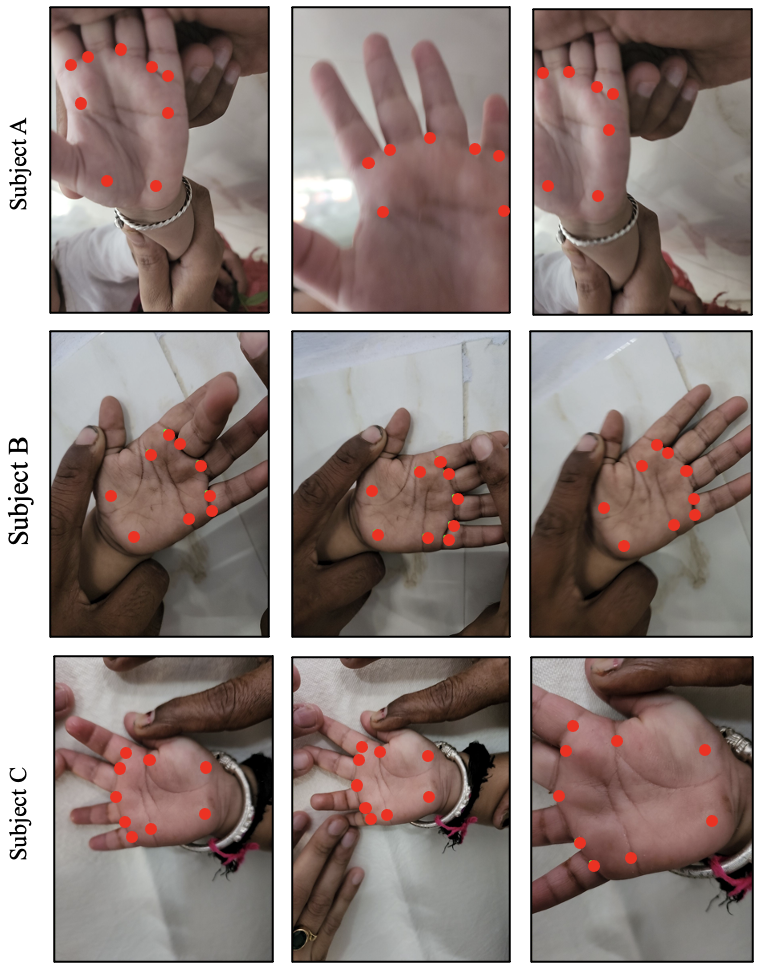}
    \caption{Sample images from three distinct identities having large intra-class variability in terms of occlusion, blurring and stand-off distance.}
    \label{fig:large_intra}
\end{figure}

Due to the continuous capture of images in PalmMobile SDK Android application coupled with the uncooperative nature of the subjects results in a large intra-class variability. This includes blurring, half-open fists and occlusions from the operator’s hand and other background elements due to the unexpected movement of the subjects. Fig. \ref{fig:large_intra} shows examples of 3 distinct palms having images with larger intra-class variability. It is in these cases where the number of misclassified samples increases.

\end{document}